\documentclass[runningheads]{llncs}
\usepackage{graphicx}
\usepackage{booktabs}

\usepackage{amsmath}
\usepackage{multirow}
\usepackage[usenames,dvipsnames,table]{xcolor}
\usepackage{pifont}
\usepackage{hyperref}
\usepackage{adjustbox}
\usepackage{tablefootnote}

\hypersetup{
    colorlinks=true,
    linkcolor=blue,
    filecolor=magenta,      
    citecolor=blue,
    urlcolor=blue,
    }

\newcommand{\lsim}{\mathcal{L}_{\text{sim}}} 
\newcommand\imarksubject{\raisebox{-0.2em}{\includegraphics[width=1.3em]{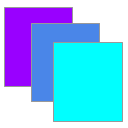}}}
\newcommand\imarkuniversal{\raisebox{-0.2em}{\includegraphics[width=1.3em]{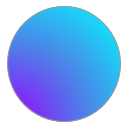}}}

\newcommand{\cmark}{\ding{51}}%
\newcommand{\xmark}{\ding{55}}%

\newcommand{\gcmark}[0]{{\textcolor{ForestGreen}{\cmark}}}
\newcommand{\rxmark}[0]{{\textcolor{BrickRed}{\xmark}}}

\begin{document}
\title{\texttt{uniGradICON}: A Foundation Model for \\ Medical Image Registration}
%
%
\author{
    Lin Tian\inst{1} \and
    Hastings Greer\inst{1} \and
    Roland Kwitt \inst{2} \and 
    Francois-Xavier Vialard \inst{3}  \and \\
    Raul San Jose Estepar \inst{4} \and
    Sylvain Bouix \inst{5} \and
    Richard Rushmore \inst{6} \and \\
    Marc Niethammer\inst{1}}

\institute{University of North Carolina at Chapel Hill \and
University of Salzburg, Austria \and
University Paris-Est \and
Brigham and Women's Hospital \and
ÉTS Montréal \and
Boston University
}
\authorrunning{Tian et al.}
%
%
\maketitle              
\begin{abstract}
Conventional medical image registration approaches directly optimize over the parameters of a transformation model. These approaches have been highly successful and are used generically for registrations of different anatomical regions. Recent deep registration networks are incredibly fast and accurate but are only trained for specific tasks. Hence, they are no longer generic registration approaches. We therefore propose \texttt{uniGradICON}, a first step toward a foundation model for registration providing 1) great performance \emph{across} multiple datasets which is not feasible for current learning-based registration methods, 2) zero-shot capabilities for new registration tasks suitable for different acquisitions, anatomical regions, and modalities compared to the training dataset, and 3) a strong initialization for finetuning on out-of-distribution registration tasks. \texttt{UniGradICON} unifies the speed and accuracy benefits of learning-based registration algorithms with the generic applicability of conventional non-deep-learning approaches. We extensively trained and evaluated \texttt{uniGradICON} on twelve different public datasets. Our code and the \texttt{uniGradICON} model are available at \url{https://github.com/uncbiag/uniGradICON}.

\end{abstract}

\begin{figure}[t]
    \centering
    \includegraphics[width=0.95\textwidth]{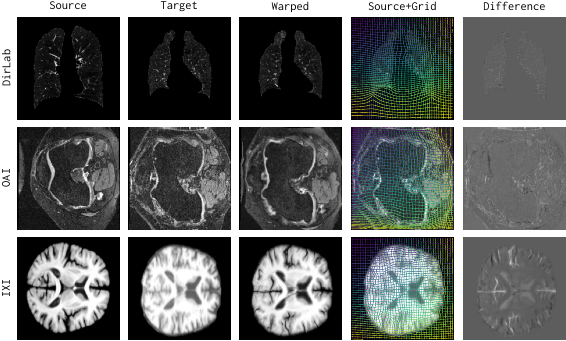}
    \caption{Example \texttt{uniGradICON} registrations. Prediction only w/o IO.}
    \label{fig:enter-label}
    \vspace{-0.5cm}
\end{figure}
\section{Introduction}
\label{section:introduction}

Conventional registration methods~\cite{avants2008symmetric,klein2009elastix,modat2010fast,heinrich2012globally} directly estimate spatial correspondences for an image pair. They can be used for a wide variety of registration tasks, can be highly accurate, but are often slow as they estimate registration parameters from scratch for every registration pair by numerical optimization. More recent supervised~\cite{yang2017quicksilver,cao2017deformable,sokooti2017nonrigid} and unsupervised~\cite{de2017end,balakrishnan2019voxelmorph} learning-based registration approaches \emph{predict} spatial correspondences much faster using a deep registration network. These learning-based approaches have achieved significant accuracy improvements by advanced transformation models~\cite{shen2019networks,niethammer2019metric,niethammer2019metric}, network structures~\cite{mok2020fast,chen2022transmorph}, training schemes~\cite{hering2019mlvirnet,de2019deep,mok2020large}, and similarity measures~\cite{tian2023same++,mok2024modality}. Learning-based methods now outperform conventional methods on various registration tasks in terms of accuracy and speed. However, \emph{current learning-based methods require training task-specific networks} making them much less flexible than approaches that use numerical optimization per image pair. 

Our \textbf{key question} is: Can we train a \emph{universal} registration network that can be used as generically as conventional registration algorithms while retaining the speed and accuracy advantages of learning-based, but task-specific, methods? One can imagine obtaining such a model by training \emph{one} registration network over many different datasets to obtain a \emph{universal foundation model} for registration. However, this is not straightforward. The crux is that  picking the right kind of registration hyperparameters (for regularizer and similarity measure) is important for good registration performance for conventional and learning-based registration alike. However, while this tuning for conventional approaches might be tedious, learning-based approaches generally require the costly training of a new network instance. While some work to adapt hyperparameters after training exists for deep registration networks~\cite{hoopes2021hypermorph}, the  fundamental issue is that \emph{different} registration tasks generally require \emph{different} hyperparameters. However, a universal registration network will be trained using only one \emph{fixed} set of hyperparameters, which then would be suboptimal for some registration tasks.

\vskip0.5ex
Recent work~\cite{greer2021icon,tian2023gradicon} on learning-based registration allows training task-specific registration networks \emph{using exactly the same training procedure and hyperparameters}. This is made possible~\cite{tian2023gradicon} by replacing conventional regularizers (e.g., diffusion) by gradient inverse consistency (\texttt{GradICON}) regularization. While conventional regularizers need to be carefully balanced with an image similarity measure the \texttt{GradICON} regularizer is weaker as it only encourages invertibility of the transform. 
This weaker regularity allows the network to discover what transformations are supported by the data and thereby facilitates training with the same hyperparameters across different datasets. Hence, a key question we explore is if \texttt{GradICON} regularization can also be used to train a universal foundation model for registration. 

\vskip1ex
\noindent
\textbf{Contributions.} 1) We develop (to the best of our knowledge) the first foundation model for registration; 2) we show that our \texttt{uniGradICON} model can achieve excellent registration accuracy across multiple datasets; 3) we demonstrate that \texttt{uniGradICON} can be successfully used to register images from different image sources, anatomical regions, and image modalities.

\section{Material \& method}
\subsection{Dataset curation and pre-processing}
\textbf{Dataset.} We created a \emph{composite training dataset} from publicly available medical image corpora. This composite dataset contains various anatomical regions (e.g., lung, knee, brain, and abdomen), different modalities (e.g., CT, CBCT, and MRI), and various deformation patterns (e.g., lung inspiration/expiration or inter-subject anatomical mappings). See Tab.~\ref{tab:datasets} in the appendix for details.

\vskip1ex
\noindent
\textbf{Intensity pre-processing.} For CT images, we clip the Hounsfield Units (HU) to $[-1000, 1000]$ and then linearly normalize to $[0,1]$. For MRIs, we clip the maximum intensity at the 99th percentile and then similarily standardize to $[0,1]$. \emph{This pre-processing is the same across the training and test phases}.

\vskip1ex
\noindent
\textbf{Spacing pre-processing.} We resize all images to $[175,175,175]$ using trilinear interpolation. Thus, image spacing of the network input images may not be isotropic. 
We always evaluate on the original images by linearly interpolating the output transformation fields back to the original spacing.

\subsection{Registration network}
We rely on the publicly-available \texttt{GradICON} registration network~\cite{tian2023gradicon} which uses a \emph{two-step} registration process: 1) images are run through a three-level multi-resolution registration network where, at each level, a UNet accepts the warped image from the previous level and the target image. In addition, the input images are downsampled to $\frac{1}{4},\frac{1}{2},1$ for each level, respectively; 2) images are run through one UNet that accepts the warped image from the first step and the target image at full resolution. All four UNets have the same architecture.

\subsection{Training protocol and experimental setting}
\textbf{Training dataset.} Our composite dataset (see Tab.~\ref{tab:datasets} in the appendix) for training contains intra- and inter-patient data. The intra-patient dataset (dataset 1) contains 899 pairs of inspiration/expiration lung CT images. The inter-patient datasets contain 2532, 1076, and 30 images. We randomly sample two images within each dataset, leading to 3,205,512, 578,888, and 450 possible distinct image pairs for datasets 2-4, respectively. To prevent bias due to  the differing numbers of paired images for the different datasets, we randomly sample $N=1000$ image pairs from each dataset during each epoch, resulting in 4000 3D image pairs per epoch.

\vskip1ex
\noindent
\textbf{Training loss.}
The loss proposed in \texttt{GradICON}~\cite{tian2023gradicon} has the following formulation:
\begin{equation}
	\label{equ:loss_function}
		\mathcal{L} = \lsim\left(I^A \circ \Phi^{AB}, I^B\right) + \lsim\left(I^B \circ \Phi^{BA}, I^A\right) + 
		\lambda \lVert \nabla \left(\Phi^{AB} \circ \Phi^{BA}\right) - \mathbf{I}\rVert_F^2\enspace.
\end{equation}
Given an ordered image pair $(I^A,I^B)$, the registration network outputs the transformation map $\Phi^{AB}$ which maps $I^A$ to the space of $I^B$. By swapping the input pair $(I^B, I^A)$, we obtain the estimated inverse map $\Phi^{BA}$. The similarity loss $\mathcal{L}_{\text{sim}}$ is computed between the warped image $I^A \circ \Phi^{AB}$ and the target image $I^B$, and vice versa. We use localized normalized cross correlation ($1-\text{LNCC}$) as similarity measure. The third term in Eq.~\eqref{equ:loss_function} is the gradient inverse consistency regularizer, which penalizes differences between the Jacobian of the composition of $\Phi^{AB}$ with $\Phi^{BA}$ and the identity matrix $\mathbf{I}$;  $\lVert\cdot\rVert_F^2$ is the Frobenius norm, $\lambda>0$.

\vskip1ex
We use \texttt{GradICON} as the basic building block of our approach because it provides excellent registration performance for a variety of datasets \emph{using exactly the same hyperparameter and training settings}~\cite{tian2023gradicon}. We expect this behavior to allow training a better registration model \emph{across} our composite dataset compared to using competing approaches that rely on task-specific training and hyperparameter settings. Sec.~\ref{sec:experiments} empirically supports this hypothesis.

\vskip1ex
\noindent
\textbf{Training hyperparameters.} We train the first step for 800 epochs and the second for 200 epochs with a learning rate of 5e-5 and a balancing constant of $\lambda=1.5$. These are the \emph{default settings} for \texttt{GradICON}\footnote{Better performance might be achievable by further hyperparameter tuning. But this is not the focus of the current study which is targeted at establishing if it is feasible to train a deep registration network with good performance across multiple datasets.}.

\subsection{Model availability and development plan}
We will release the source code and the  \texttt{uniGradICON} model. We will periodically update our released model to include more anatomical regions, modalities, and deformation patterns. We appreciate suggestions for  datasets to be included.

\section{Experiments}
\label{sec:experiments}

\texttt{UniGradICON}'s contributions are as follows. \textbf{First}, it obtains state-of-the-art (SOTA) or close-to SOTA accuracy \emph{without retraining}, resulting in similar generality as conventional registration approaches. Hence, our model bridges the gap between the versatility of conventional optimization-based registration algorithms (e.g., \texttt{ANTs}~\cite{avants2008symmetric}, \texttt{Elastix}~\cite{klein2009elastix}, or \texttt{NifyReg}~\cite{modat2010fast}, which while largely motivated by brain registration are general purpose registration tools) and the speed and accuracy of task-specific deep registration networks. \textbf{Second}, it can provide a satisfying baseline for zero-shot out-of-distribution registrations. \textbf{Third}, when combined with finetuning on an out-of-distribution dataset, it can provide on-par performance to task-specific registration networks for that dataset. To evaluate \texttt{uniGradICON} w.r.t. these three aspects, {\bf we test on} {\bf 1)} in-distribution test datasets (Sec.~\ref{sec:exp_in_distribution}), {\bf 2)} out-of-distribution datasets with zero-shot inference (Sec.~\ref{sec:exp_out_distribution_zero_shot}), and {\bf 3)} by finetuning on an out-of-distribution dataset (Sec.~\ref{sec:exp_out_distribution_finetune}).

\subsection{Performance on in-distribution tasks}
\label{sec:exp_in_distribution}
To test the in-distribution performance of \texttt{uniGradICON}, we evaluate our model on datasets 5-7 (Tab.~\ref{tab:datasets}) and the validation set of dataset 4 (L2R-Abdomen). 

\vskip1ex
Tab.~\ref{tab:exp_in_distribution} shows that \texttt{uniGradICON} achieves on par or comparable performance to models trained \emph{specifically} for a dataset. We further observe that \texttt{uniGradICON} 1) generalizes much better across datsets than a task-specific model (\texttt{GradICON}-lung), 2) consistently outperforms an excellent conventional registration approach (\texttt{SyN}), and 3) performs significantly better than a \texttt{VoxelMorph}-based foundation model (Tab.~\ref{tab:implementation_setting} in the appendix), also trained on the composite dataset. These results verify our hypothesis that the weaker \texttt{GradICON} regularizer of \texttt{uniGradICON} indeed allows successfully training \emph{one} universal registration model compared to the lower accuracy of a diffusion-regularizer-based \texttt{VoxelMorph} variant. We did not succeed in training a \texttt{LapIRN}-based~\cite{mok2020large} foundation model, likely due to the need for task-specific hyperparameter tuning. Also, we note that the universal \texttt{VoxelMorph} model may benefit from affine pre-alignment which is not needed for \texttt{uniGradICON}. However, due to the large number of training samples for a universal model, affine alignment would have to rely on on-the-fly affine registration. This in turn would require a universal affine registration network which does not yet exist.

\begin{table}[t]
    \centering
    \begin{adjustbox}{width=\columnwidth,center}
    \begin{tabular}{llcccccccc} \toprule
  && \multicolumn{2}{c}{COPDGene}& \multicolumn{2}{c}{OAI}& \multicolumn{2}{c}{HCP}& \multicolumn{2}{c}{L2R-Abdomen}\\ \midrule
          &&  \textbf{mTRE}[mm]&  $\%|J|_{<0}$&  \textbf{DICE}&  $\%|J|_{<0}$&  \textbf{DICE}&  $\%|J|_{<0}$&  \textbf{DICE}& $\%|J|_{<0}$\\
  &Initial& 23.36& & 7.6& & 53.4& & 25.9&\\ \midrule
          \multirow{10}{*}{\imarksubject}&\texttt{VM}-lung~\cite{balakrishnan2019voxelmorph}&  9.88~\cite{tian2023gradicon}&  0.0&  -&  -&  -&  -&  -& -\\
          &\texttt{GradICON}-lung~\cite{tian2023gradicon}&  1.93~\cite{tian2023gradicon}&  3e-4&  38.0&  7.9e-4&  73.0& 4.4e-5&  18.1& 3.4e-2\\
          &\texttt{GradICON}-lung(IO)~\cite{tian2023gradicon}&  1.31~\cite{tian2023gradicon}&  2e-4&  70.0&  8.0e-3&  79.3&  1.9e-4&  36.9& 6.6e-1\\
          &\texttt{LapIRN}~\cite{mok2020large}& 2.92~\cite{tian2023gradicon}& 0.0& - & - & -& -& -&\\
          &\texttt{VM+Affine}-knee~\cite{balakrishnan2019voxelmorph}&  -&  -&  66.1~\cite{tian2023gradicon}&  1.3e-3&  -&  -&  -& -\\
          &\texttt{VM}-knee~\cite{balakrishnan2019voxelmorph}&  -&  -&  46.1~\cite{tian2023gradicon}&  2.8e-3&  -&  -&  -& -\\
          &\texttt{GradICON}-knee~\cite{tian2023gradicon}&  -&  -&  70.1~\cite{tian2023gradicon}&  2.61e-2&  -&  -&  -& -\\
          &\texttt{GradICON}-knee(IO)~\cite{tian2023gradicon}&  -&  -&  71.2~\cite{tian2023gradicon}&  4.20e-3&  -&  -&  -& -\\
          &\texttt{GradICON}-brain~\cite{tian2023gradicon}&  -&  -&  -&  -&  78.7~\cite{tian2023gradicon}&  1.2e-3&  -& -\\ 
          &\texttt{GradICON}-brain(IO)~\cite{tian2023gradicon}&  -&  -&  -&  -&  80.5~\cite{tian2023gradicon}&  4e-4&  -& -\\ \midrule
          \multirow{5}{*}{\imarkuniversal}&\texttt{SyN}~\cite{avants2008symmetric}& 1.79~\cite{tian2023gradicon}& -& 65.7~\cite{tian2023gradicon}& 0& 75.8~\cite{tian2023gradicon}&0 & 25.2 & 0\\ 
          &\texttt{VoxelMorph}-SVF~\cite{balakrishnan2019voxelmorph}  & 19.21 & 0 &  55.0&  1.1e-4&  44.2& 1.7e-2 &  33.8&   7.2e-2\\
          &\texttt{uniGradICON}&  2.26&  9.3e-5&  68.9&  3.9e-2&  76.2&  6.4e-5& 48.3&  3.1e-1\\
          &\texttt{uniGradICON}(IO)&  1.40&  9.0e-5&  70.3&  2.2e-2&  78.9&  2.2e-4&  52.2& 9.6e-1  \\ \bottomrule
    \end{tabular}
    \end{adjustbox}
    \vskip1ex 
    \caption{Comparison between task-specific (\raisebox{-0.3em}{\includegraphics[width=1.2em]{Figures/ICON_subject_model.png}}) and universal 
    (\raisebox{-0.3em}{\includegraphics[width=1.2em]{Figures/ICON_universal_model.png}}) models based on \texttt{VoxelMorph}, \texttt{LapIRN} and \texttt{uniGradICON}. References indicate a published result.}
    \label{tab:exp_in_distribution}
    \vspace{-0.5cm}
\end{table}

\begin{table}[htp]
    \centering
    \begin{tabular}{ccccc} \toprule
 & Anatom. region&~Deformation~& ~Acquisition~&~Modality~\\ \midrule
         In-distribution&  \gcmark&  \gcmark&  \gcmark&  \gcmark\\
         Out-distribution (\textbf{Type 1})&  \gcmark&  -&  \rxmark&  \gcmark\\ 
 Out-distribution (\textbf{Type 2})& \rxmark& \rxmark& -&\gcmark\\
 Out-distribution (\textbf{Type 3})& \gcmark& -& -&\rxmark\\ \bottomrule
    \end{tabular}
    \vskip1.5ex
    \caption{Types of generalization. \gcmark and \rxmark\ denote whether corresponding data have been included in the composite training dataset. $-$ denotes a data type that we do not strictly test \texttt{uniGradICON}'s generalization capability on.}
    \label{tab:generalization_type}
    \vspace{-0.5cm}
\end{table}

\subsection{Performance on out-of-distribution tasks}\label{sec:exp_out_distribution_zero_shot}

We evaluate the zero-shot performance of \texttt{uniGradICON} on out-of-distribution datasets. We classify out-of-distribution datasets into three categories: {\bf Type 1} comprises datasets that contain the same anatomical regions as the composite training dataset but originate from different sources (e.g., comparing HCP to IXI); {\bf Type 2} are datasets with unseen anatomical regions not covered by the composite dataset; {\bf Type 3} are datasets of modalities not contained in the composite training dataset. Tab.~\ref{tab:generalization_type} provides an overview of these different types.

\begin{table}[htp]
    \centering
    \begin{adjustbox}{width=\columnwidth,center}
    \begin{tabular}{lcccccccc} \toprule
 & \multicolumn{3}{c}{L2R-NLST}& \multicolumn{2}{c}{L2R-OASIS} && \multicolumn{2}{c}{IXI}\\
 & \multicolumn{2}{c}{Validation}& Test& \multicolumn{2}{c}{Validation}&Test& \multicolumn{2}{c}{Test}\\ \midrule
         &  \textbf{mTRE}[mm]&  $\%|J|_{<0}$ & \textbf{mTRE}[mm]&  \textbf{DICE}&  $\%|J|_{<0}$ &\textbf{DICE}&  \textbf{DICE}& $\%|J|_{<0}$\\
 Initial& 10.22&  & 11.2& 57.18&  &56& 40.6 &\\ \midrule
 \texttt{Learn2Reg}~\cite{hering2022learn2reg} Top-1& -& -& \cellcolor{blue!10}1.44& -&  -&\cellcolor{blue!10}82& -&-\\
 \texttt{Learn2Reg}~\cite{hering2022learn2reg} Top-5& -& -& \cellcolor{blue!10}2.04& -& -& \cellcolor{blue!10}78& -&-\\ \midrule
 \texttt{VoxelMorph}~\cite{balakrishnan2019voxelmorph}& -& - & -& -& - &-& 73.2~\cite{chen2022transmorph}&1.522\\
 \texttt{TransMorph}~\cite{chen2022transmorph}& -& - & -& -& - &-& 75.4~\cite{chen2022transmorph}&1.579\\ \midrule
         \texttt{SyN}& \cellcolor{blue!10}3.04 & 9.8-1  & -& \cellcolor{blue!10}75.6  &  1.5e-2 &-&  64.5~\cite{chen2022transmorph}& $<$1e-4\\
         \texttt{uniGradICON}& \cellcolor{blue!10}2.07 & 4.7e-4  & -& \cellcolor{blue!10}79.0&  8.9e-4 &-&  70.6& 7.4e-3\\ 
        \texttt{uniGradICON}(IO)& \cellcolor{blue!10}1.77& 2.0e-2 & -& \cellcolor{blue!10}79.6& 1.9e-3 &-& 71.3 & 1.8e-1\\ \bottomrule
    \end{tabular}
    \end{adjustbox}
    \vskip1.5ex
    \caption{Evaluation of \texttt{uniGradICON} on {\bf Type 1} out-of-distribution tasks with zero-shot inference and instance optimization (IO).}
    \label{tab:exp_different_data_sources}
    \vspace{-0.5cm}
\end{table}

\noindent
\textbf{Different sources (Type 1).} We test zero-shot inference of \texttt{uniGradICON} on one lung dataset, L2R-NLST, and two brain datasets, L2R-OASIS and IXI. This experiment studies how \texttt{uniGradICON} generalizes to images of a modality and of anatomical regions contained in the composite dataset \emph{but} acquired as part of different studies.
As we do not have access to the test set of the \texttt{Learn2Reg} challenge, we use the validation set for testing. \emph{This is  valid because we do not train on the \texttt{Learn2Reg} dataset}. As there are no existing foundation models for registration, we compare to \texttt{SyN}. 
To provide context for how the current task-specific models perform on these datasets, we included the results reported in the \texttt{Learn2Reg} official paper and website for the top 5 methods. Tab.~\ref{tab:exp_different_data_sources} shows that \texttt{uniGradICON} performs better than \texttt{SyN} across all three registration tasks with \textbf{zero-shot inference}, with further improvements achievable by instance optimization (IO). The \texttt{uniGradICON} results are within the performance range of the top 5 \texttt{Learn2Reg} methods trained and tuned for the specific tasks. Note that while these results are not directly comparable, we assume that the validation and test sets from the same dataset share the same distribution and, hence, share the same trends. We conclude that \texttt{uniGradICON} is a strong out-of-the-box benchmark model for \textbf{Type 1} out-of-distribution tasks.

\noindent
\textbf{Different regions (Type 2).} We test the generalizability of \texttt{uniGradICON} to registrations for unseen anatomical regions. We train  \texttt{uniGradICON} excluding L2R-Abdomen from the composite dataset and test on the L2R-Abdomen validation set. This is challenging as the images and deformation patterns are both not seen during training. Tab.~\ref{tab:exp_different_regions} shows that although \texttt{uniGradICON}(wo Abdomen) increases the initial DICE score from 25.9\% to 34.1\%, we observe an accuracy drop compared to \texttt{uniGradICON} which was trained on the full composite dataset. However, \texttt{uniGradICON}(wo Abdomen) achieves a better registration result than \texttt{SyN}. It also provides a good initialization for instance optimization, mitigating most of the performance drop and coming close to the Top-5 \texttt{Learn2Reg} accuracy for a task-specific model. 
We conclude that \texttt{uniGradICON} can be a good out-of-the-box baseline for \textbf{Type 2} out-of-distribution tasks. 

\begin{table}[htp]
    \centering
        \begin{adjustbox}{width=\columnwidth,center}
    \begin{tabular}{lccccccccc} \toprule
 & \multicolumn{2}{c}{COPDGene}& \multicolumn{2}{c}{OAI}& \multicolumn{2}{c}{HCP}& \multicolumn{3}{c}{L2R-Abdomen}\\
 & & & & & & & \multicolumn{2}{c}{Validation}&Test\\ \midrule
         &  \textbf{mTRE}&  $\%|J|_{<0}$&  \textbf{DICE}&  $\%|J|_{<0}$&  \textbf{DICE}&  $\%|J|_{<0}$&  \textbf{DICE}& $\%|J|_{<0}$ &\textbf{DICE}\\
 Initial& 23.36& & 7.6& & 53.4& & 25.9& &28\\ \midrule
 \texttt{Learn2Reg}~\cite{hering2022learn2reg} Top-1& -& -& -& -& -& -& -& -&\cellcolor{blue!10}69\\
 \texttt{Learn2Reg}~\cite{hering2022learn2reg} Top-5& -& -& -& -& -& -& -& -&\cellcolor{blue!10}49\\ \midrule
        \texttt{SyN}~\cite{avants2008symmetric}& 1.79~\cite{tian2023gradicon}& -& 65.7~\cite{tian2023gradicon}& 0& 75.8~\cite{tian2023gradicon}&0 & \cellcolor{blue!10}25.2 & 0&-\\
         \texttt{uniGradICON}& 2.26&  9.3e-5&  68.9&  3.9e-2&  76.2&  6.4e-5& \cellcolor{blue!10}48.3&  3.1e-1 &-\\
 \texttt{uniGradICON}(wo Abdomen)& 2.20& 5.6e-6& 68.9& 8.9e-3& 76.8& 1.2e-5& \cellcolor{blue!10}34.1&1.9e-2 &-\\ 
 \texttt{uniGradICON}(wo Abdomen) (IO)& 1.41 &2.6e-5 &70.2 &1.2e-2 &79.0 & 1.5e-4 & \cellcolor{blue!10}45.3&7.9e-1 &-\\ \bottomrule
    \end{tabular}
    \end{adjustbox}
    \vskip1.5ex
    \caption{Evaluation of \texttt{uniGradICON} on \textbf{Type 2} out-of-distribution tasks with zero-shot inference and instance optimization (IO).}
    \label{tab:exp_different_regions}
    \vspace{-1cm}
\end{table}

\vskip0.5ex
\noindent
\textbf{Different Modalities (Type 3).} We test how \texttt{uniGradICON} generalizes when the input images have different modalities from the composite training dataset. We use the L2R-CBCT and the L2R-MRCT datasets for evaluation. The L2R-CBCT dataset contains paired images of CT and CBCT where both the CBCT and the combination of CT and CBCT are absent in the composite dataset. For the L2R-CTMRI dataset, the input combination of CT and MRI is unseen during  training. Tab.~\ref{tab:exp_different_modalities} shows that although \texttt{uniGradICON} has not been trained for multi-modal registration, its accuracy is within the range of the top 5 well-tuned and task-specific methods on L2R-CBCT, highlighting its strong generalization ability to unseen modalities and multi-modality registration problems. Compared to the excellent performance on L2R-CBCT, \texttt{uniGradICON} is not as strong as the well-tuned and trained task-specific methods on L2R-CTMRI. We hypothesize that the combination of CBCT and CT is visually closer to the CT pairs in the composite dataset than the combination of MRI and CT. Thus, it is more challenging for \texttt{uniGradICON} to generalize to the L2R-CBCT task. 
We conclude that \texttt{uniGradICON} can be used as an out-of-the-box baseline method for \textbf{Type 3} out-of-distribution tasks. 
\begin{table}[htp]
    \centering
     \begin{adjustbox}{width=0.7\columnwidth,center}
    \begin{tabular}{lcccccc} \toprule
 & \multicolumn{3}{c}{L2R-CBCT}& \multicolumn{3}{c}{L2R-CTMR}\\
 & \multicolumn{2}{c}{Validation}&Test& \multicolumn{2}{c}{Validation}&Test\\ \midrule
         & \textbf{DICE}&  $\%|J|_{<0}$ &\textbf{DICE}&  \textbf{DICE}&  $\%|J|_{<0}$ &\textbf{DICE}\\
 Initial& 31.3&  &28.0& 31.3& &33\\ \midrule
 \texttt{Learn2Reg}~\cite{hering2022learn2reg} Top-1& -& - &\cellcolor{blue!10}63.2&- & -&\cellcolor{blue!10}75\\
 \texttt{Learn2Reg}~\cite{hering2022learn2reg} Top-5& -& - &\cellcolor{blue!10}56.9&- & -&\cellcolor{blue!10}71\\ \midrule
         \texttt{SyN}~\cite{avants2008symmetric} & \cellcolor{blue!10}57.4&0 &-&\cellcolor{blue!10}45.0&0&-\\
         \texttt{uniGradICON}&  \cellcolor{blue!10}57.0& 4.7e-4  &-&  \cellcolor{blue!10}50.0& 4e-2  &-\\ 
 \texttt{uniGradICON} (IO)& \cellcolor{blue!10}59.9& 0 &-& \cellcolor{blue!10}66.8 & 6.1e-1 &-\\ \midrule
 \texttt{uniGradICON} (finetune)& \cellcolor{blue!10}60.3& 3.6e-1 &-& -&- &-\\
 \texttt{uniGradICON} (finetune+IO)& \cellcolor{blue!10}63.7&  8.9e-1&-&- &- &-\\ \bottomrule
    \end{tabular}
    \end{adjustbox}
    \vskip1.5ex
    \caption{Evaluation of \texttt{uniGradICON} on {\bf Type 3} out-of-distribution tasks with zero-shot inference, instance optimization (IO), and target task finetuning.}
    \label{tab:exp_different_modalities}
\end{table}

\subsection{Performance of finetuning on out-of-distribution dataset}
\label{sec:exp_out_distribution_finetune}
We study the performance of \texttt{uniGradICON} when used as an \emph{initialization} and finetuned on a target registration task. We test on the \textbf{Type 3} out-of-distribution dataset L2R-CBCT. We finetune \texttt{uniGradICON} with the L2R-CBCT training dataset (excluding the validation set) for 4,000 epochs with the learning rate and hyper-parameters used initially. 
Tab.~\ref{tab:exp_different_modalities} shows that the finetuned \texttt{uniGradICON} model is better than the best task-specific \texttt{Learn2Reg} model.

\section{Conclusion, limitations, and future work}
We have developed \texttt{uniGradICON}, a foundation registration model that performs on par with task-specific SOTA methods for in-distribution registration tasks (Tab.~\ref{tab:exp_in_distribution}), alleviating the burden of training new registration networks for every task. \texttt{UniGradICON} achieves comparable performance to well-trained SOTA task-specific methods on datasets collected from different sources (Tab.~\ref{tab:exp_different_data_sources}) and that contain out-of-distribution modalities (Tab.~\ref{tab:exp_different_modalities}), demonstrating \texttt{uniGradICON}'s good out-of-the-box baseline registration performance. We also showed that finetuning \texttt{uniGradICON} on an unseen target dataset can further improve accuracy.

\vskip1ex
\noindent\textbf{Limitations and future work.} \texttt{UniGradICON} currently includes limited in-distribution registration tasks. We will include more diverse training datasets in future work. Although \texttt{uniGradICON} shows multi-modal generalization abilities (cf. Tab.~\ref{tab:exp_different_modalities}), its support for multi-modal registration could likely be improved by using self-supervised modality-agnostic representations, by training on additional multi-modal image datasets, or by using $1-\text{LNCC}^2$ or normalized mutual information as the similarity measure. A self-supervised representation may also help improve the current limited zero-shot performance for unseen regions (cf. Tab.~\ref{tab:exp_different_regions}). \texttt{UniGradICON} only uses images: further improvements might also be possible by including segmentations for training and instance optimization. Finally, we remark that \texttt{uniGradICON} uses \emph{the} \texttt{GradICON} deep network; using a larger network most likely improves performance.

\subsection{Acknowledgements}

This work was supported by NIH grants 1R01AR072013, 1R01AR082684, \\ 1R01EB028283, 1R21MH132982, RF1MH126732, 1R01HL149877,\\ 
5R21LM013670, and R01NS125307. The work expresses the views of the authors, not of NIH. Roland Kwitt was supported in part by the Land Salzburg within the EXDIGIT project 20204-WISS/263/6-6022 and projects 0102-F1901166- KZP,\\ 
20204-WISS/225/197-2019. Sylvain Bouix was supported in part by Natural Sciences and Engineering Research Council grants RGPIN-2023-05443 and CRC-2022-00183. The knee imaging data were obtained from the controlled access datasets distributed from the Osteoarthritis Initiative (OAI), a data repository housed within the NIMH Data Archive. OAI is a collaborative informatics system created by NIMH and NIAMS to provide a worldwide resource for biomarker identification, scientific investigation and OA drug development. Dataset identifier: NIMH Data Archive Collection ID: 2343. The brain imaging data were provided by the Human Connectome Project, WU-Minn Consortium (Principal Investigators: David Van Essen and Kamil Ugurbil; 1U54MH091657) funded by the 16 NIH Institutes and Centers that support the NIH Blueprint for Neuroscience Research; and by the McDonnell Center for Systems Neuroscience at Washington University. The lung imaging data was provided by the COPDGene study. Further data was provided by the Learn2Reg challenge as well as through IXI (Information eXtraction from Images -- EPSRC GR/S21533/02).

\newpage
\bibliographystyle{splncs04}
\bibliography{reference}

\appendix
\clearpage

\section{Dataset and implementation details}

\begin{table}[htp]
    \small
    \begin{center}
    \begin{adjustbox}{width=\columnwidth,center}
    \begin{tabular}{lcccccc}\toprule
         \textbf{Dataset}&  Anatom. & \# of & \# per &   \# of &Type& Modality\\
        & region& patients& patient&  pairs&&\\ \midrule
 1. COPDGene~\cite{regan2011genetic}&   Lung&899&  2&   899&Intra-pat.& CT\\
 2. OAI~\cite{nevitt2006osteoarthritis}&   Knee&2532&  1&   3,205,512&Inter-pat.& MRI\\
 3. HCP~\cite{van2012human}& Brain& 1076& 1&  578,888&Inter-pat.&MRI\\ 
 4. L2R-Abdomen~\cite{xu2016evaluation}& Abdomen& 30 & 1 &  450&Inter-pat. &CT\\ \midrule
 5. Dirlab-COPDGene~\cite{castillo2013reference}&   Lung&10&  2&   10&Intra-pat.& CT\\
 6. OAI-test~\cite{nevitt2006osteoarthritis}& Knee& 301& 1&  301&Inter-pat.&MRI\\
 7. HCP-test~\cite{van2012human}& Brain& 32& 1&  100&Inter-pat.&MRI\\
 8. L2R-NLST-val\tablefootnote{\url{https://www.cancerimagingarchive.net/collection/nlst/}}~\cite{NLST2011,Clark2013}& Lung& 10& 2&  10&Intra-pat.&CT\\
 9. L2R-OASIS-val~\cite{marcus2007open,hoopes2021hypermorph}& Brain& 20& 1&  19&Inter-pat.&MRI\\
 10. IXI-test\tablefootnote{\url{https://brain-development.org/ixi-dataset/}}& Brain& 115& 1&  115&Atlas-pat.&MRI\\
 11. L2R-CBCT-val~\cite{CBCT2016,Hugo2017}& Lung& 3& 3&  6&Intra-pat.&CT/CBCT\\
 12. L2R-CTMR-val~\cite{clark2013cancer,akin2016radiology,CTMRI_KIRP,CTMRI_LIHC}& Abdomen& 3& 2&  3&Intra-pat.&CT/MRI\\\midrule
 13. L2R-CBCT-train~\cite{CBCT2016,Hugo2017}& Lung& 3& 11&  22&Intra-pat.&CT/CBCT\\\bottomrule
    \end{tabular}
    \end{adjustbox}
    \end{center}
    \caption[datasets]{\label{tab:datasets} Listing of datasets used for training and evaluation. Datasets 1-4 are used for training \texttt{uniGradICON}.  Datasets 5-7 are used for in-distribution evaluation. Datasets 8-13 are used for out-of-distribution zero-shot evaluation and assessment of fine-tuning results. We follow the official \texttt{Learn2Reg} (https://learn2reg.grand-challenge.org/) dataset split so that our results can be evaluated on the official website. According to this split, the images in the L2R-Abdomen validation set are included in the training set.}
\vspace{-1cm}
\end{table}

\begin{table}[]
    \centering
    \begin{tabular}{lccc} \toprule
         \textbf{Methods} &  Transformation&Similarity &Official default \\
 & Models& Measure&hyper-parameters\\ \midrule
         \texttt{SyN}~\cite{avants2008symmetric}& 
     Affine + SVF&Mutual Information &\gcmark\\
 \texttt{VoxelMorph}~\cite{balakrishnan2019voxelmorph}& SVF&Mean Squared Error &\gcmark\\
 \texttt{LapIRN}~\cite{mok2020large}& SVF& LNCC&\gcmark\\
 \texttt{uniGradICON}& DVF&LNCC &\gcmark\\ \bottomrule
 \end{tabular} 
 \vskip1.5ex
    \caption{Settings of the methods in the experiments. We train the universal \texttt{VoxelMorph} using MSE instead of LNCC because the model faces difficulty converging properly when using LNCC. \texttt{VoxelMorph} requires an affine pre-alignment. However, for a large training dataset, it is not feasible to compute the pre-alignment without a universal affine registration network.}
    \label{tab:implementation_setting}
\vspace{-1cm}
\end{table}

\section{Some example registration results}
\begin{figure}
    \centering
    \includegraphics[height=0.95\textheight]{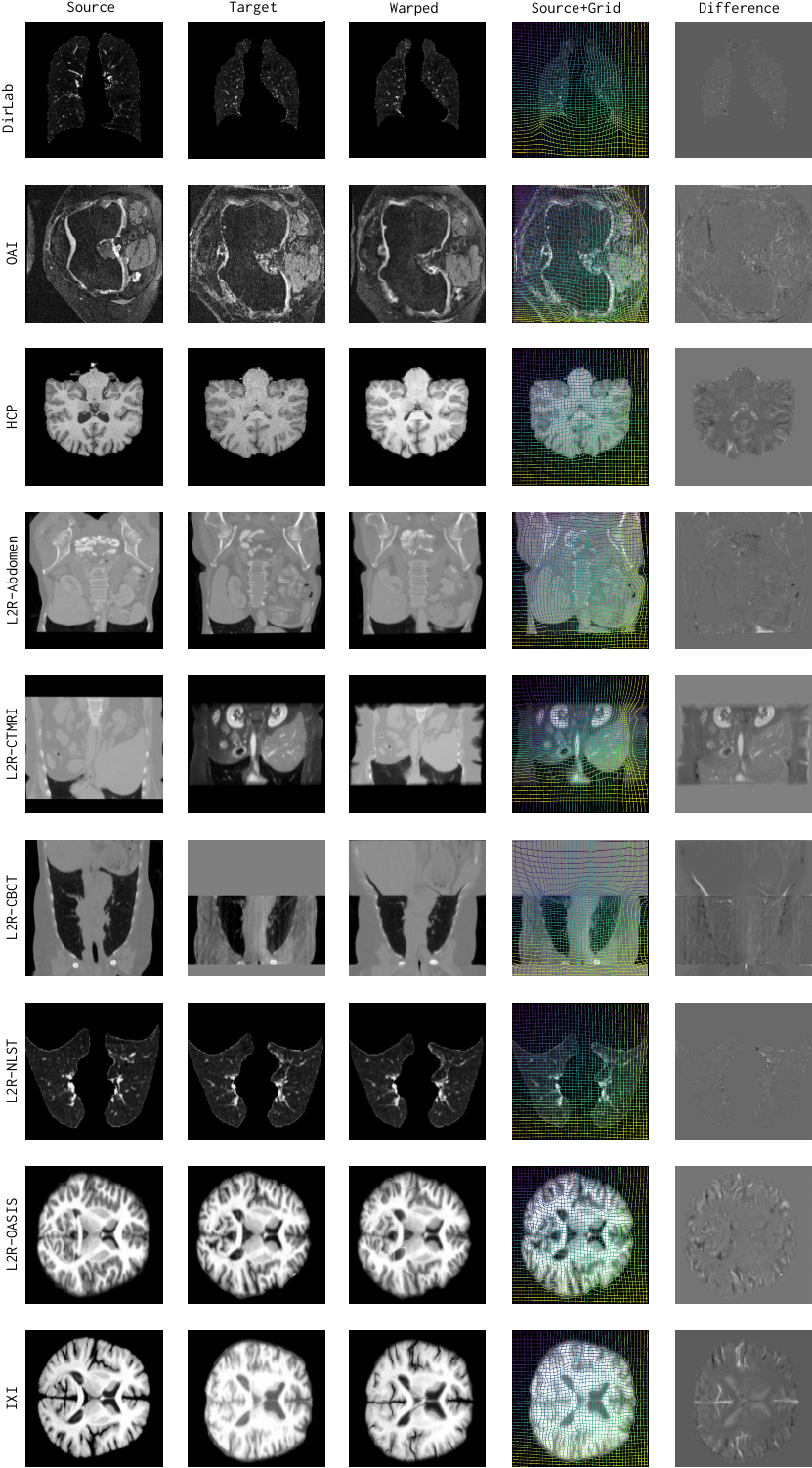}
    \caption{Visualization of \texttt{uniGradICON} registration results for \emph{zero-shot inference}. We display images as they are presented to our \texttt{uniGradICON} foundation model, i.e., not necessarily based on the typical anatomical convention. Note that while for a task-specific network it is important to use consistent image orientations it is much less clear that this is the case for a universal registration network, as such a network is ideally expected to be able to handle images of any orientation. In future work this could be further studied, for example, by exploring specific orientation-changing data augmentation strategies.}
    \label{fig:registration_visualization}
\end{figure}

Fig.~\ref{fig:registration_visualization} shows example \texttt{uniGradICON} registration results.

\end{document}